\documentclass[letterpaper]{article} 
\usepackage{aaai25}  
\usepackage{times}  
\usepackage{amsmath}
\usepackage{amssymb}
\usepackage{helvet}  
\usepackage{courier}  
\usepackage[hyphens]{url}  
\usepackage{graphicx} 
\usepackage{natbib}  
\usepackage{caption} 
\usepackage{booktabs}
\usepackage{algorithm}
\usepackage{algorithmic}
\usepackage{multirow}

\usepackage{newfloat}
\usepackage{listings}
\DeclareCaptionStyle{ruled}{labelfont=normalfont,labelsep=colon,strut=off} 
\floatstyle{ruled}
\newfloat{listing}{tb}{lst}{}
\floatname{listing}{Listing}
\title{MPTSNet: Integrating Multiscale Periodic Local Patterns and Global Dependencies for Multivariate Time Series Classification}
\author{
    Yang Mu\textsuperscript{\rm 1},
    Muhammad Shahzad\textsuperscript{\rm 1},
    Xiao Xiang Zhu\textsuperscript{\rm 1, 2}\thanks{Corresponding author}
}
\affiliations{
    \textsuperscript{\rm 1}Technical University of Munich, Germany\\
    \textsuperscript{\rm 2}Munich Center for Machine Learning (MCML), Germany

    \text{\{yang.mu, muhammad.shahzad, xiaoxiang.zhu\}}@tum.de
}

\usepackage{bibentry}

\begin{document}

\maketitle

\begin{abstract}
Multivariate Time Series Classification (MTSC) is crucial in extensive practical applications, such as environmental monitoring, medical EEG analysis, and action recognition. Real-world time series datasets typically exhibit complex dynamics. To capture this complexity, RNN-based, CNN-based, Transformer-based, and hybrid models have been proposed. Unfortunately, current deep learning-based methods often neglect the simultaneous construction of local features and global dependencies at different time scales, lacking sufficient feature extraction capabilities to achieve satisfactory classification accuracy. To address these challenges, we propose a novel Multiscale Periodic Time Series Network (MPTSNet), which integrates multiscale local patterns and global correlations to fully exploit the inherent information in time series. Recognizing the multi-periodicity and complex variable correlations in time series, we use the Fourier transform to extract primary periods, enabling us to decompose data into multiscale periodic segments. Leveraging the inherent strengths of CNN and attention mechanism, we introduce the PeriodicBlock, which adaptively captures local patterns and global dependencies while offering enhanced interpretability through attention integration across different periodic scales. The experiments on UEA benchmark datasets demonstrate that the proposed MPTSNet outperforms 21 existing advanced baselines in the MTSC tasks.
\end{abstract}

%
\begin{links}
\link{Code}{https://github.com/MUYang99/MPTSNet}
\link{Datasets}{https://timeseriesclassification.com/dataset.php}
\end{links}

\section{Introduction}
Multivariate time series classification (MTSC) has garnered significant attention due to its widespread applications across various domains. This classification technique plays a crucial role in analyzing complex, multi-dimensional temporal data, offering insights that are invaluable in fields ranging from healthcare~\cite{wang2022systematic, an2023comprehensive}, human activity recognition~\cite{yang2015deep, li2023human}, traffic~\cite{zhao2024metarocketc} to environmental monitoring~\cite{ienco2007tiselac, russwurm2023end} and industrial processes~\cite{farahani2023time}.

\begin{figure}[t]
\centering
\includegraphics[width=0.85\columnwidth]{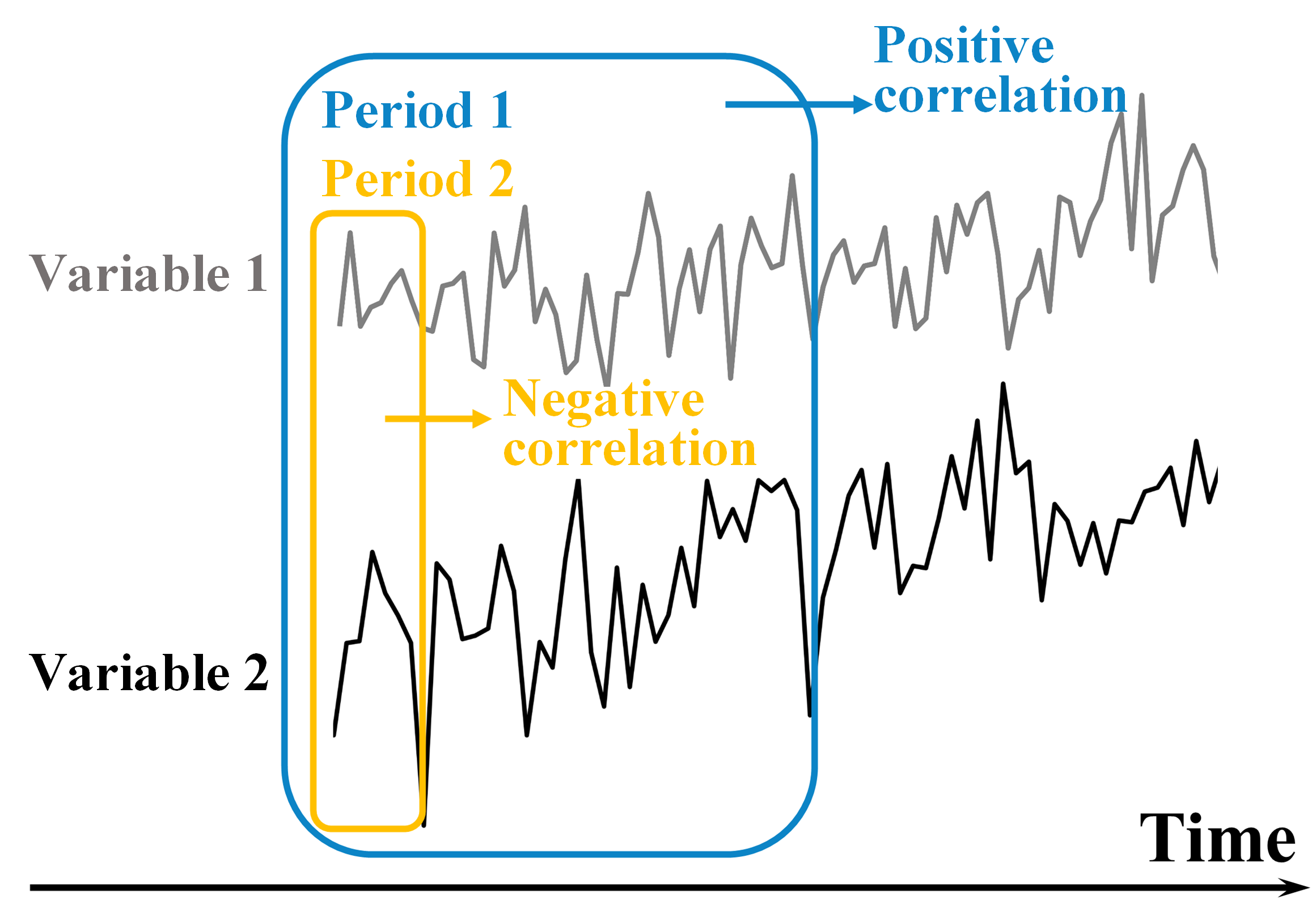}
\caption{Varying correlations between the subsequences of two variables across different periodic scales in MTS data.}
\label{dcorr}
\end{figure}

The challenges in solving MTSC tasks stem from two primary factors. Firstly, time series data inherently possesses periodic characteristics, presenting similar trends across different periods, which leads to data redundancy. Simultaneously, hidden information overlaps and interacts across multiple periods~\cite{wu2023timesnet}, complicating the exploration of time series data. Secondly, time series subsequences among different variables exhibit varying degrees of correlation or even opposite correlations at different periodic scales~\cite{cai2024msgnet} in MTS data, as shown in Fig.~\ref{dcorr}. Thus, the extraction of local intra-period features and the capture of global inter-period dependencies at multiple periodic scales become paramount for effective MTSC.

Traditional approaches to time series classification, such as bag-of-patterns~\cite{baydogan2013bag} or shapelet-based methods~\cite{lines2012shapelet}, typically require a preprocessing step to transform time series into a wide range of subsequences or patterns as candidate features. Subsequently, discriminative subsequences are selected from these candidates for classification purposes. However, this approach often results in an expansive feature space, which not only complicates the feature selection process but may also lead to decreased accuracy~\cite{schafer2017multivariate}, particularly in multivariate settings where the complexity of data is inherently higher.

Recently, deep neural networks have demonstrated remarkable potential in various time series tasks, outperforming traditional methods in many aspects. RNN-based methods~\cite{karim2017lstm, yu2021analysis} utilize loops within their architecture to maintain and propagate information across time steps, but limit the ability to capture long-term dependencies due to the vanishing gradient problem. CNN-based methods~\cite{cui2016multi, zhao2017convolutional} excel at learning spatial hierarchies of features through convolutional filters, the inductive bias allows them to capture local patterns effectively, but they are less adept at modeling global features comprehensively. Transformer-based methods~\cite{wen2022transformers, zuo2023svp} leverage self-attention mechanisms to model dependencies between different positions in the input sequence, adeptly handling long-range dependencies. However, they struggle to effectively extract features from adjacent time points that exhibit localized pattern characteristics. While these models have shown promise, they often fall short in simultaneously constructing local features and global dependencies within time series data across different periodic scales, resulting in suboptimal feature extraction.

\begin{figure}[t]
\centering
\includegraphics[width=0.99\columnwidth]{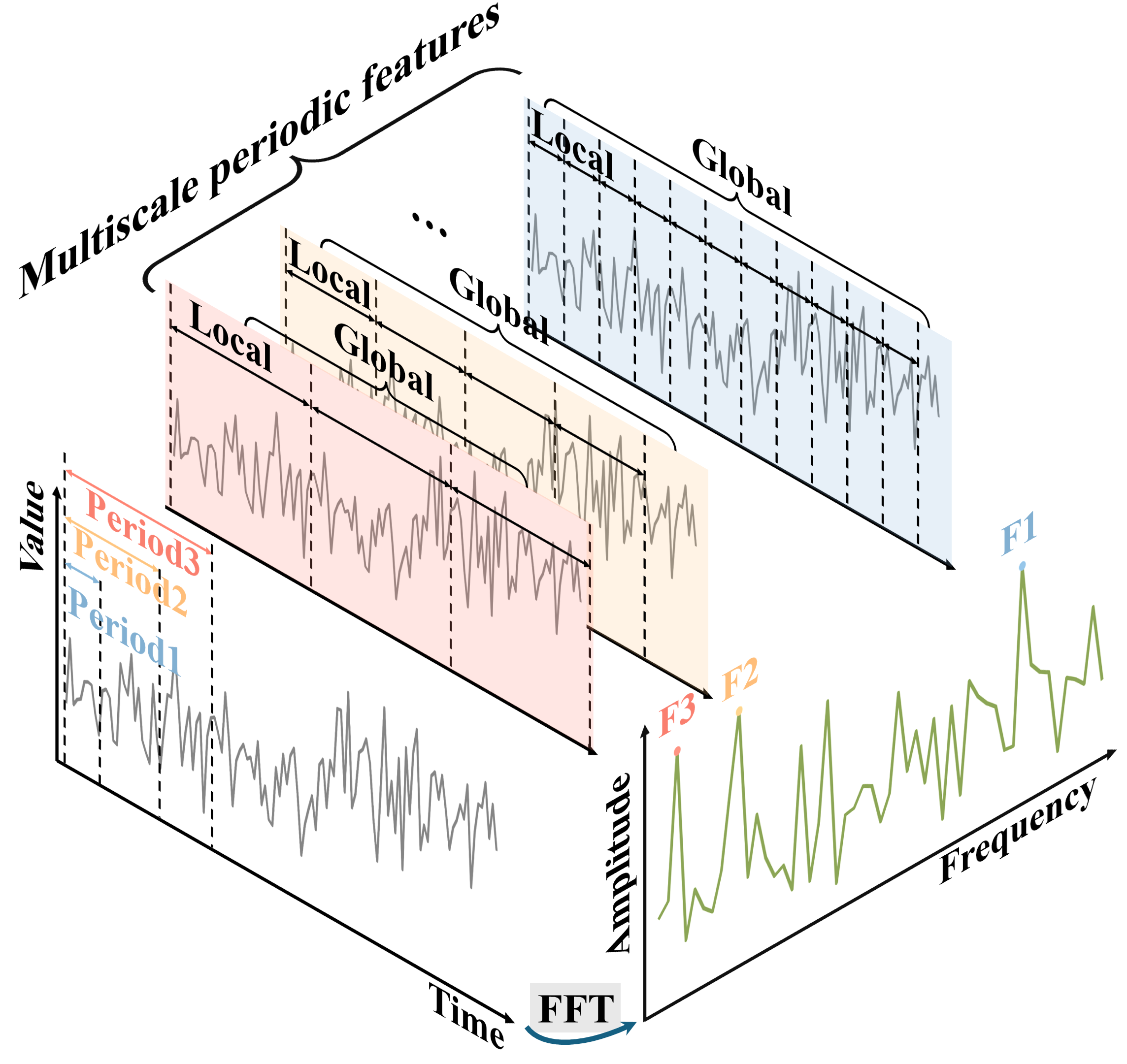} 
\caption{Multiscale periodic feature analysis of time series data using FFT. By decomposing time series into multiple frequency components, it reveals both local patterns and global dependencies across different periodic scales.}
\label{coreidea}
\end{figure}

To address these limitations, we propose a general framework called Multiscale Periodic Time Series Network (MPTSNet). The core concept, illustrated in Fig.~\ref{coreidea}, involves transforming time-domain data into frequency-domain representations using the Fast Fourier Transform (FFT). This approach decomposes complex time series into multiple frequency components, revealing both local intra-period patterns and global inter-period dependencies across various periodic scales. The architecture of MPTSNet is supported by the PeriodicBlock, which decomposes time series into multiscale periodic segments. These segments are then processed by two key components sequentially: an Inception-based Local Extractor, which captures corresponding local features, and an Attention-based Global Capturer, which extracts global dependencies. This modular architecture enables MPTSNet to effectively analyze time series data at multiple periodic scales. Our contributions can be summarized in three key aspects:
\begin{itemize}
    \item We propose an end-to-end general framework MPTSNet that addresses the multi-periodicity of time series and complex correlations among variable subsequences at different time scales. This novel architecture is capable of extracting multiscale periodic features effectively.
    \item Inspired by the strengths of CNN in capturing local features and attention mechanism in modeling global dependencies, we design a PeriodicBlock comprising an Inception-based Local Extractor and an Attention-based Global Capturer to extract both local features and global dependencies from time series data.
    \item Our extensive experiments on real-world datasets demonstrate that the proposed MPTSNet outperforms existing advanced baselines in MTSC tasks. Furthermore, MPTSNet offers enhanced interpretability through its ability to localize temporal patterns with integrated global attention across different periodic scales.
\end{itemize}

\section{Related Work}
\subsection{Multivariate Time Series Classification}
Recent years have witnessed the emergence of various deep learning models for MTSC. These can be broadly categorized into three main groups: 1) CNN-based models. The Multichannel Deep Convolutional Neural Network (MCDCNN)~\cite{zheng2014time} pioneered the application of CNNs to MTSC by capturing intra-variable features through one-dimensional convolutions and combining them with fully connected layers. More recently, OS-CNN~\cite{tangomni} has been proposed, featuring an innovative Omni-Scale block that uses a set of prime numbers as convolution kernel sizes to effectively cover receptive fields across all scales, optimizing performance across different datasets. 2) RNN/CNN Hybrid models. Models such as LSTM-FCN~\cite{karim2017lstm} and MLSTM-FCN~\cite{karim2019multivariate} have been introduced, combining LSTM layers to capture short- and long-term dependencies with stacked CNN layers to extract features from the time series. 3) Transformer-based models. These models have gained prominence in \textit{General Time Series Analysis}, addressing multiple tasks including classification, imputation, short-term and long-term forecasting, and anomaly detection~\cite{chen2024timemil}. Notable examples include FEDformer~\cite{zhou2022fedformer}, Flowformer~\cite{wu2022flowformer}, PatchTST~\cite{nie2023a}, and Crossformer~\cite{zhang2023crossformer}, which have been developed and refined in recent years, leveraging their excellent scaling behaviors. Additionally, MLP-based~\cite{zhang2022less, li2023mts}, GNN-based~\cite{liu2024todynet} and MIL-based~\cite{chen2024timemil} models have also achieved impressive performance. Since MIL-based models employ a binary paradigm for multi-class problems, unlike other approaches, they are excluded from quantitative comparisons.

However, a key assumption in many of these models is that the correlations between different variables remain constant across various time resolutions. This assumption often leads to inadequate representation of pairwise relationships between variables at different periodic scales.

\begin{figure*}[t]
\centering
\includegraphics[width=\textwidth]{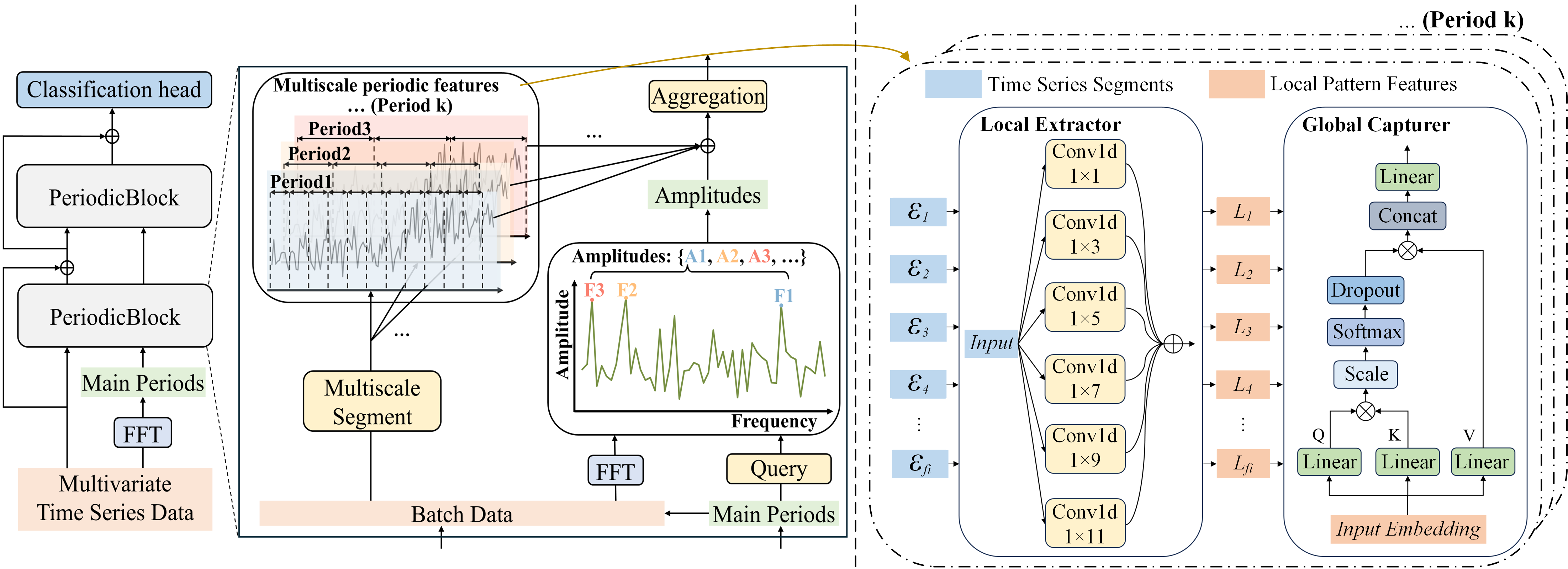}
\caption{Overview of the Multiscale Periodic Time Series Network (MPTSNet). The model processes multivariate time series data through FFT to identify the main periods. The data is then segmented into multiscale periodic components, where Local Extractor and Global Capturer in PeriodicBlock extract features sequentially at different scales. The features are aggregated by corresponding amplitudes and passed through the Classification head for final prediction.}
\label{ov}
\end{figure*}

\subsection{Local and Global Modeling for Sequence}
Modeling local patterns and global dependencies has proven effective for correlation learning and feature extraction in sequence modeling~\cite{wang2023micn}. In speech recognition, the SOTA model Interformer~\cite{lai2023interformer} employs convolution to extract local information and self-attention to capture long-term dependencies in parallel branches. However, this parallel processing of local and global features might introduce redundancy, potentially reducing efficiency. In general time series analysis, TimesNet~\cite{wu2023timesnet} transforms 1D time series into various 2D tensors using CNN-based layers to capture both local and global variations. Directly capturing local and global features with CNNs may lead to a loss in precision. Additionally, while TimesNet provides visualizations of the 2D transformations, interpreting these in the context of the original 1D time series can be challenging. In time series forecasting, MSGNet~\cite{cai2024msgnet} employs a graph convolution module for inter-series correlation learning and a multi-head attention module for intra-series correlation learning. However, it introduces increased complexity and computational cost due to its adaptive graph convolution module; MICN~\cite{wang2023micn} combines local feature extraction and global correlation modeling using a convolution-based structure; however, its performance can be highly sensitive to the choice of hyperparameters, such as scale sizes in the multi-scale convolution, requiring meticulous tuning for optimal results. Moreover, the use of full CNN layers in MICN sacrifices the interpretability. 

To address these existing limitations, we propose MPTSNet, which sequentially applies convolutional and attention modules to capture local and global features. By requiring only the number of extraction scales, other parameters are adaptively configured based on time series characteristics, reducing sensitivity to parameter tuning and avoiding redundant computations. Moreover, this approach enhances the interpretability of MTSC by integrating multi-scale attention maps.

\section{Methodology}
\subsection{Problem Formulation}
MTSC involves the analysis of a set of time series data, where each series is composed of multiple variables observed over time. A MTS data is formally defined as \( X = \{x_1, x_2, \ldots, x_d\} \in \mathbb{R}^{d \times l} \), where \( d \) is the number of variables and \( l \) is the length of the series. In the context of MTSC, the goal is to classify these time series into predefined categories. Given a dataset \( \mathcal{X} = \{X_1, X_2, \ldots, X_m\} \in \mathbb{R}^{m \times d \times l} \) with corresponding labels \( \eta = \{y_1, y_2, \ldots, y_m\} \), where \( m \) denotes the number of time series, the task is to train a classifier \( f(\cdot) \) capable of predicting the label \( y \) for new, unlabeled MTS data. 

\subsection{Model Architecture Overview}
The overview of our proposed MPTSNet is shown in Fig.~\ref{ov}, which is designed to extract local patterns and global dependencies at various periodic scales. Initially, the model identifies the main periods from the entire training dataset using FFT. The core component, PeriodicBlock, processes time series data in parallel. It begins by segmenting batch data into multiple scales based on the identified main periods. Subsequently, the Local Extractor and Global Capturer are employed within the PeriodicBlock to extract intra-period local and inter-period global features sequentially, as depicted in the right panel of Fig.~\ref{ov}. To account for potential discrepancies in frequency-amplitude relationships between batch data and the overall training set, FFT is also applied to each batch. This process queries the amplitudes corresponding to the main periods, which are then used as weights calculated by Softmax to aggregate multiscale periodic features. The PeriodicBlock outputs a weighted sum of features extracted from different periodic scales. Residual connections between stacked PeriodicBlocks enhance information flow. Finally, the extracted features are fed into the Classification head to generate predicted labels.

\subsection{Main Periods Identification}
Grounded in the inherent characteristics of time series data, we aim to enhance MTSC accuracy by exploiting two fundamental properties: multi-periodicity and complex correlations among variables across different periodic scales, as mentioned before. The identification of main periods is a critical step in our methodology, as it serves as the foundation for subsequent multiscale analysis.

Inspired by the work~\cite{wu2023timesnet}, we employ the FFT to identify the principal periods. This technique effectively converts time-domain data into the frequency domain, allowing for a comprehensive analysis of the underlying periodic structures. The process can be formalized as follows:

{\small
\begin{equation}
\mathbf{AMP} = \text{Avg}(\text{Amplitudes}(\text{FFT}(X \in \mathbb{R}^{d \times l}))),
\end{equation}
}

where $\ X \in \mathbb{R}^{d \times l}$ represents the input time series, $d$ denotes the number of variables, and $l$ is the sequence length. The function $\text{FFT}(\cdot)$ computes the Fast Fourier Transform, while $\text{Amplitudes}(\cdot)$ calculates the amplitude values. $\text{Avg}(\cdot)$ averages the amplitudes across all variables, resulting in $\mathbf{AMP} \in \mathbb{R}^l$, which represents the mean amplitude for each frequency.

{\small
\begin{equation}
f_1, \dots, f_k = \mathop{\arg \operatorname{Topk}}\limits_{f_* \in \{1, \dots, \lfloor \frac{T}{2} \rfloor \}} (\mathbf{AMP}),
\end{equation}
}

{\small
\begin{equation}
\quad p_i = {l} / {f_i}, \quad i \in \{1, \ldots, k\},
\end{equation}
}

To focus on the most significant periodic components, we select the top $k$ amplitudes using the $\mathop{\arg \operatorname{Topk}}$ function. This operation yields the frequencies $f_1, \ldots, f_k$, which correspond to the $k$ most prominent periods $p_1, \ldots, p_k$. The parameter $k$ is a hyperparameter that determines the number of periodic scales to consider in subsequent analyses.

The identification of these main periods establishes the groundwork for multiscale analysis. This approach allows us to decompose the original time series into multiple periodic components, and such decomposition is crucial for understanding the complex temporal dynamics present in MTS data.

The extracted periods serve as the basis for reshaping the input data into multiple representations, each corresponding to a different time scale. This transformation can be expressed as:

{\small
\begin{equation}
{X'}^i = \text{Reshape}_{p_i,f_i}(\text{Pad}(X)), \quad i \in \{1, \ldots, k\}.
\end{equation}
}

where $\text{Pad}(\cdot)$ extends the time series with zeros to ensure compatibility with the reshaping operation, and ${X'}^i \in \mathbb{R}^{d \times p_i \times f_i}$ represents the $i$-th reshaped time series based on the $i$-th periodic scale.


\subsection{PeriodicBlock}

The PeriodicBlock serves as the core component of our MPTSNet architecture, designed to extract both local and global features from the reshaped time series data at different periodic scales. Prior to processing by the PeriodicBlock, each reshaped tensor ${X'}^i \in \mathbb{R}^{d \times p_i \times f_i}$ is embedded into a higher-dimensional space:

{\small
\begin{equation}
\mathcal{E'}^i = \text{Embedding}({X'}^i), \quad i \in \{1, \ldots, k\}.
\end{equation}
}

where $\mathcal{E'}^i \in \mathbb{R}^{d_\text{embed} \times p_i \times f_i}$ and $d_\text{embed}$ is the embedding dimension.

\subsubsection{Local Extractor}

To capture specific and comprehensive intra-period local patterns, we propose a multiscale adaptive convolution module inspired by the Inception architecture. For each of the $f_i$ embeddings $\mathcal{E}^i \in \mathbb{R}^{d_\text{embed} \times p_i}$, this module comprises parallel 1D convolutional layers with varying kernel sizes:

{\small
\begin{equation}
L_j^{i, ker} = \text{Conv1D}_{ker}(\mathcal{E}^i), \quad ker \in \{1, 3, 5, 7, 9, 11\},
\end{equation}
}

where $\text{Conv1D}_{ker}$ represents a 1D convolutional layer with kernel size $ker$. The outputs of these convolutions are concatenated along the channel dimension:

{\small
\begin{equation}
L_j^i = \text{Mean}(L_j^{i, ker}), \quad j \in \{1, \ldots, f_i\}.
\end{equation}
}

This multi-kernel size Inception-based approach allows the model to capture local patterns at various granularities, enhancing its ability to detect relevant features.

\subsubsection{Global Capturer}

To model inter-period dependencies and adapt to different time steps for multiscale periods, we designed a Global Capturer based on the multi-head attention mechanism. The Global Capturer transforms each local feature $L^i$ into query, key, and value representations:

{\small
\begin{equation}
Q^i = W_Q L^i, \quad K^i = W_K L^i, \quad V^i = W_V L^i,
\end{equation}
}

where $W_Q$, $W_K$, and $W_V$ are learnable weight matrices. The multi-head attention is then computed as:

{\small
\begin{equation}
\text{Attention}(Q^i, K^i, V^i) = \text{Softmax}\left(\frac{Q^i (K^i)^T}{\sqrt{d_k}}\right) V^i,
\end{equation}
}

{\small
\begin{equation}
G^i = \text{Concat}(\text{head}_1, \ldots, \text{head}_h)W^O.
\end{equation}
}

where $h$ is the number of attention heads, $d_k$ is the dimension of each head, and $W^O$ is a learnable output projection matrix. This mechanism allows the model to capture complex global dependencies across all time series segments.

\begin{table*}[t]
\centering
\small
\setlength{\tabcolsep}{1mm}
\begin{tabular}{l|cccccccccccc}
\toprule
\multirow{2}{*}{Data/Model} & LSTNet & LSSL & FEDf. & Flowf. & SCINet & Dlinear & PatchTST & MICN & TimesNet & Crossf. & M.TCN & \multirow{2}{*}{Ours} \\
& \scriptsize \textit{SIG.'18} & \scriptsize \textit{ICLR'22} & \scriptsize \textit{ICLR'22} & \scriptsize \textit{ICLR'22} & \scriptsize \textit{NIPS'22} & \scriptsize \textit{AAAI'23} & \scriptsize \textit{ICLR'23} & \scriptsize \textit{ICLR'23} & \scriptsize \textit{ICLR'23} & \scriptsize \textit{ICLR'23} & \scriptsize \textit{ICLR'24} &  \\
\midrule
EthanolConcentration & \underline{39.9} & 31.1 & 31.2 & 33.8 & 34.4 & 36.2 & 32.8 & 35.3 & 35.7 & 38.0 & 36.3 & \textbf{43.3} \\
FaceDetection & 65.7 & 66.7 & 66.0 & 67.6 & 68.9 & 68.0 & 68.3 & 65.2 & 68.6 & 68.7 & \textbf{70.8} & \underline{69.8} \\
Handwriting & 25.8 & 24.6 & 28.0 & \underline{33.8} & 23.6 & 27.0 & 29.6 & 25.5 & 32.1 & 28.8 & 30.6 & \textbf{34.4} \\
Heartbeat & 77.1 & 72.7 & 73.7 & \underline{77.6} & 77.5 & 75.1 & 74.9 & 74.7 & \textbf{78.0} & \underline{77.6} & 77.2 & 75.6 \\
JapaneseVowels & 98.1 & 98.4 & 98.4 & \underline{98.9} & 96.0 & 96.2 & 97.5 & 94.6 & 98.4 & \textbf{99.1} & 98.8 & 98.6 \\
PEMS-SF & 86.7 & 86.1 & 80.9 & 86.0 & 83.8 & 75.1 & 89.3 & 85.5 & \underline{89.6} & 85.9 & 89.1 & \textbf{94.2} \\
SelfRegulationSCP1 & 84.0 & 90.8 & 88.7 & 92.5 & 92.5 & 87.3 & 90.7 & 86.0 & 91.8 & 92.1 & \textbf{93.4} & \underline{92.8} \\
SelfRegulationSCP2 & 52.8 & 52.2 & 54.4 & 56.1 & 57.2 & 50.5 & 57.8 & 53.6 & 57.2 & \underline{58.3} & \textbf{60.3} & 57.2 \\
SpokenArabicDigits & \textbf{100} & \textbf{100} & \textbf{100} & 98.8 & 98.1 & 81.4 & 98.3 & 97.1 & 99.0 & 97.9 & 98.7 & \underline{99.5} \\
UWaveGestureLibrary & \underline{87.8} & 85.9 & 85.3 & 86.6 & 85.1 & 82.1 & 85.8 & 82.8 & 85.3 & 85.3 & 86.7 & \textbf{88.1} \\
\midrule
Ours 1-to-1-Wins & 3 & 4 & 9 & 3 & 3 & 8 & 1 & 10 & 5 & 7 & 5 & - \\
Ours 1-to-1-Draws & 1 & 2 & 0 & 1 & 0 & 0 & 0 & 0 & 1 & 0 & 0 & - \\
Ours 1-to-1-Losses & 6 & 4 & 1 & 6 & 7 & 2 & 9 & 0 & 4 & 3 & 5 & - \\
\midrule
Avg. accuracy ($\uparrow$) & 71.8 & 70.9 & 70.7 & 73.2 & 71.7 & 67.9 & 72.5 & 70.0 & 73.6 & 73.2 & \underline{74.2} & \textbf{75.4} \\
Avg. rank ($\downarrow$) & 6 & 7.5 & 7.6 & 4.6 & 6.7 & 8.6 & 6.1 & 9.2 & 4.1 & 4.5 & \underline{3} & \textbf{2.4}
 \\
\bottomrule
\end{tabular}%
\caption{Setting 1 experimental results. Performance comparison with the recent advanced \textit{General Time Series Analysis} frameworks on 10 UEA datasets. The best results are highlighted in \textit{bold}, while the second best are \textit{underlined}.}
\label{tab:set1}
\end{table*}

\subsubsection{Adaptive Aggregation and Classification}

To fuse the $k$ multiscale periodic features, we employ an adaptive aggregation using weights calculated from the queried amplitudes:

{\small
\begin{equation}
\alpha_i = \text{Softmax}(\mathbf{AMP}_i), \quad i \in \{1, \ldots, k\},
\end{equation}
}

where $\mathbf{AMP}_i$ represents the amplitude corresponding to the $i$-th selected frequency. The aggregated feature representation is then computed as:

{\small
\begin{equation}
Z = \sum_{i=1}^k \alpha_i G^i.
\end{equation}
}

This adaptive weighting scheme ensures that the model emphasizes the most relevant periodic components in the final representation.

The aggregated features $Z$ are then passed through the Classification head, typically consisting of fully connected layers followed by a softmax activation:

{\small
\begin{equation}
y = \text{Softmax}(W_c Z + b_c),
\end{equation}
}

where $W_c$ and $b_c$ are the weight matrix and bias vector of the classification layer, respectively, and $y$ is the predicted class probability distribution.

By combining the Local Extractor for intra-period local pattern extraction, the Global Capturer for modeling inter-period global dependencies, and the adaptive aggregation mechanism, the PeriodicBlock effectively captures the complex temporal dynamics present in multivariate time series data across multiple periodic scales. 

\section{Experiments}
\subsection{Datasets}
The public UEA benchmark datasets~\cite{bagnall2018uea}, collected from various real-world applications, constitute a comprehensive archive for multivariate time series classification across several domains, including human activity, speech, medical EEG, and audio data, etc. We utilize the UEA benchmark datasets to evaluate the performance of the proposed MPTSNet. These datasets vary in length, dimensions, and the size of their training/testing sets.

\begin{table*}[t]
\centering
\small
\setlength{\tabcolsep}{1mm}
\begin{tabular}{l|ccccccccccc}
\toprule
\multirow{2}{*}{Data/Model} & \multirow{2}{*}{EDI} & \multirow{2}{*}{DTWI} & \multirow{2}{*}{DTWD} & W.+MUSE & M.-FCN & TapNet & ShapeNet & OS-CNN & MOS-CNN & TodyNet & \multirow{2}{*}{Ours} \\
&   &   &   & \scriptsize \textit{arxiv'17} & \scriptsize \textit{Neur.'19} & \scriptsize \textit{AAAI'20} & \scriptsize \textit{AAAI'21} & \scriptsize \textit{ICLR'22} & \scriptsize \textit{ICLR'22} & \scriptsize \textit{Info.'24} &  \\
\midrule
ArticularyWordRecognition & 97.0 & 98.0 & 98.7 & \underline{99.0} & 97.3 & 98.7 & 98.7 & 98.8 & \textbf{99.1} & 98.7 & 97.7 \\ 
AtrialFibrillation & 26.7 & 26.7 & 20.0 & 33.3 & 26.7 & 33.3 & 40.0 & 23.3 & 18.3 & \underline{46.7} & \textbf{53.3} \\ 
BasicMotions & 67.5 & \textbf{100} & 97.5 & \textbf{100} & 95.0 & \textbf{100} & \textbf{100} & \textbf{100} & \textbf{100} & \textbf{100} & \textbf{100} \\ 
Cricket & 94.4 & 98.6 & \textbf{100} & \textbf{100} & 91.7 & 95.8 & 98.6 & \underline{99.3} & 99.0 & \textbf{100} & 94.4 \\ 
DuckDuckGeese & 27.5 & 55.0 & 60.0 & 57.5 & 67.5 & 57.5 & \textbf{72.5} & 54.0 & 61.5 & 58.0 & \underline{68.0} \\ 
Epilepsy & 66.7 & 97.8 & 96.4 & \textbf{100} & 76.1 & 97.1 & 98.7 & 98.0 & \underline{99.6} & 97.1 & 97.1 \\ 
EthanolConcentration & 29.3 & 30.4 & 32.3 & 13.3 & 37.3 & 32.3 & 31.2 & 24.0 & \underline{41.5} & 35.0 & \textbf{43.3} \\ 
ERing & 13.3 & 13.3 & 13.3 & 43.0 & 13.3 & 13.3 & 13.3 & 88.1 & \underline{91.5} & \underline{91.5} & \textbf{94.4} \\ 
FaceDetection & 51.9 & 51.3 & 52.9 & 54.5 & 54.5 & 55.6 & 60.2 & 57.5 & 59.7 & \underline{62.7} & \textbf{69.8} \\ 
FingerMovements & 55.0 & 52.0 & 53.0 & 49.0 & 58.0 & 53.0 & 58.9 & 56.8 & 56.8 & \textbf{67.6} & \underline{64.0} \\ 
HandMovementDirection & 27.9 & 30.6 & 23.1 & 36.5 & 36.5 & 37.8 & 33.8 & 44.3 & 36.1 & \textbf{64.9} & \underline{63.5} \\ 
Handwriting & 37.1 & 50.9 & 60.7 & 60.5 & 28.6 & 35.7 & 45.1 & \textbf{66.8} & \textbf{67.7} & 43.6 & 34.4 \\ 
Heartbeat & 62.0 & 65.9 & 71.7 & 72.7 & 66.3 & \underline{75.1} & \textbf{75.6} & 48.9 & 60.4 & \textbf{75.6} & \textbf{75.6} \\ 
Libras & 83.3 & 89.4 & 87.2 & 87.8 & 85.6 & 85.0 & 85.6 & \underline{95.0} & \textbf{96.5} & 85.0 & 87.2 \\ 
LSST & 45.6 & 57.5 & 55.1 & 59.0 & 37.3 & 56.8 & 59.0 & 41.3 & 52.1 & \textbf{61.5} & \underline{60.4} \\ 
MotorImagery & 51.0 & 39.0 & 50.0 & 50.0 & 51.0 & 59.0 & 61.0 & 53.5 & 51.5 & \underline{64.0} & \textbf{65.0} \\ 
NATOPS & 86.0 & 85.0 & 88.3 & 87.0 & 88.9 & 93.9 & 88.3 & \underline{96.8} & 95.1 & \textbf{97.2} & 94.4 \\ 
PenDigits & 97.3 & 93.9 & 97.7 & 94.8 & 97.8 & 98.0 & 97.7 & 98.5 & 98.3 & \underline{98.7} & \textbf{98.9} \\ 
PEMS-SF & 70.5 & 73.4 & 71.1 & N/A & 69.9 & 75.1 & 75.1 & 76.0 & 76.4 & \underline{78.0} & \textbf{94.2} \\ 
PhonemeSpectra & 10.4 & 15.1 & 15.1 & 19.0 & 11.0 & 17.5 & \underline{29.8} & 29.9 & 29.5 & \textbf{30.9} & 14.4 \\ 
RacketSports & 86.8 & 84.2 & 80.3 & \textbf{93.4} & 80.3 & 86.8 & 88.2 & 87.7 & \underline{92.9} & 80.3 & 87.5 \\ 
SelfRegulationSCP1 & 77.1 & 76.5 & 77.5 & 71.0 & 87.4 & 65.2 & 78.2 & 83.5 & 82.9 & \underline{89.8} & \textbf{92.8} \\ 
SelfRegulationSCP2 & 48.3 & 53.3 & 53.9 & 46.0 & 47.2 & 55.0 & \textbf{57.8} & 53.2 & 51.0 & 55.0 & \underline{57.2} \\ 
StandWalkJump & 20.0 & 33.3 & 20.0 & 33.3 & 6.7 & 40.0 & \textbf{53.3} & 38.3 & 38.3 & \underline{46.7} & \textbf{53.3} \\ 
UWaveGestureLibrary & 88.1 & 86.9 & 90.3 & 91.6 & 89.1 & 89.4 & 90.6 & \textbf{92.7} & \underline{92.6} & 85.0 & 88.1 \\ 
\midrule
Ours 1-to-1-Wins & 15 & 13 & 14 & 10 & 19 & 18 & 12 & 12 & 15 & 14 & - \\
Ours 1-to-1-Draws & 2 & 2 & 2 & 2 & 1 & 4 & 1 & 3 & 2 & 3 & - \\
Ours 1-to-1-Losses & 8 & 10 & 9 & 13 & 5 & 3 & 12 & 10 & 8 & 8 & - \\
\midrule
Avg. accuracy ($\uparrow$) & 56.8 & 62.3 & 62.6 & 62.1 & 60.0 & 64.3 & 67.6 & 68.2 & 69.9 & \underline{72.6} & \textbf{74.0} \\ 
Avg. rank ($\downarrow$) & 7.52 & 6.48 & 5.8 & 5.36 & 6.4 & 5.24 & 3.84 & 4.28 & 3.8 & \underline{3.16} & \textbf{3.12} \\
\bottomrule
\end{tabular}%
\caption{Setting 2 experimental results. Performance comparison with the recent advanced MTSC-dedicated models on 25 UEA datasets. In the table, 'N/A' indicates that the results for the corresponding method could not be obtained due to memory or computational limitations~\cite{liu2024todynet}.}
\label{tab:set2}
\end{table*}

\subsection{Baselines}
\subsubsection{Setting 1 Experiments}
Based on the experimental results of papers~\cite{luo2024moderntcn, wu2023timesnet}, this set of experiments is conducted on 10 UEA datasets and compared with several recent advanced \textit{General Time Series Analysis} frameworks, which address tasks including classification, imputation, forecasting, and anomaly detection. The comparison includes ModernTCN~\cite{luo2024moderntcn}, Crossformer~\cite{zhang2023crossformer}, TimesNet~\cite{wu2023timesnet}, MICN~\cite{wang2023micn}, PatchTST~\cite{nie2023a}, Dlinear~\cite{zeng2023transformers}, SCINet~\cite{liu2022scinet}, Flowformer~\cite{wu2022flowformer}, FEDformer~\cite{zhou2022fedformer}, LSSL~\cite{gu2021efficiently}, and LSTNet~\cite{lai2018modeling}.

\subsubsection{Setting 2 Experiments}
Following the results of papers~\cite{liu2024todynet, li2021shapenet}, this set of experiments is conducted on 25 UEA datasets and compared with recent advanced MTSC-dedicated models, including TodyNet~\cite{liu2024todynet}, OS-CNN and MOS-CNN~\cite{tangomni}, ShapeNet~\cite{li2021shapenet}, TapNet~\cite{zhang2020tapnet},  MLSTM-FCN~\cite{karim2019multivariate}, and WEASEL+MUSE~\cite{schafer2017multivariate}. The traditional methods EDI, DTWI, and DTWD based on Euclidean Distance, dynamic time warping, and the nearest neighbor classifier are also included.

\subsection{Implementation Details}
The benchmark results for all baseline methods are sourced from their respective publications, ensuring consistent training parameters across comparisons. Our proposed model is implemented using a computational infrastructure consisting of a server running Ubuntu 20.04.3 LTS, equipped with 8 NVIDIA GeForce RTX 3090 GPUs. The performance is evaluated by computing the accuracy, 1-to-1 Wins/Draws/Losses, average accuracy, and average rank.

\subsection{Experimental Results}
\subsubsection{Setting 1 Results}

Table~\ref{tab:set1} presents the performance comparison of our proposed method with 11 advanced \textit{General Time Series Analysis} frameworks across 10 diverse MTSC datasets. Our approach demonstrates superior performance, achieving an average accuracy of 75.4\%, and ranks first in terms of average rank, outperforming all competitors. A detailed analysis reveals that our method consistently excels on challenging datasets. For instance, on the EthanolConcentration and PEMS-SF datasets, known for their complex patterns and high dimensionality, our method significantly outperforms other approaches. In direct comparison to the recently state-of-the-art ModernTCN, our method achieves a substantial improvement of 1.2\% in average accuracy and 0.6 in average rank, underscoring its effectiveness in handling complex multivariate time series data.

\subsubsection{Setting 2 Results}
Our method consistently demonstrates superior performance across the 25 UEA datasets compared to 10 advanced MTSC-dedicated models in Table~\ref{tab:set2}. It secures a solid win-loss record, achieving the highest average accuracy of 74\% and the lowest average rank of 3.12. Notably, our approach surpasses competitors in eleven of the twenty-five datasets. Moreover, it demonstrates strong generalization capabilities by yielding competitive results across diverse domains, including human activity recognition, healthcare, and speech recognition.

\begin{figure*}[t]
\centering
\includegraphics[width=\textwidth]{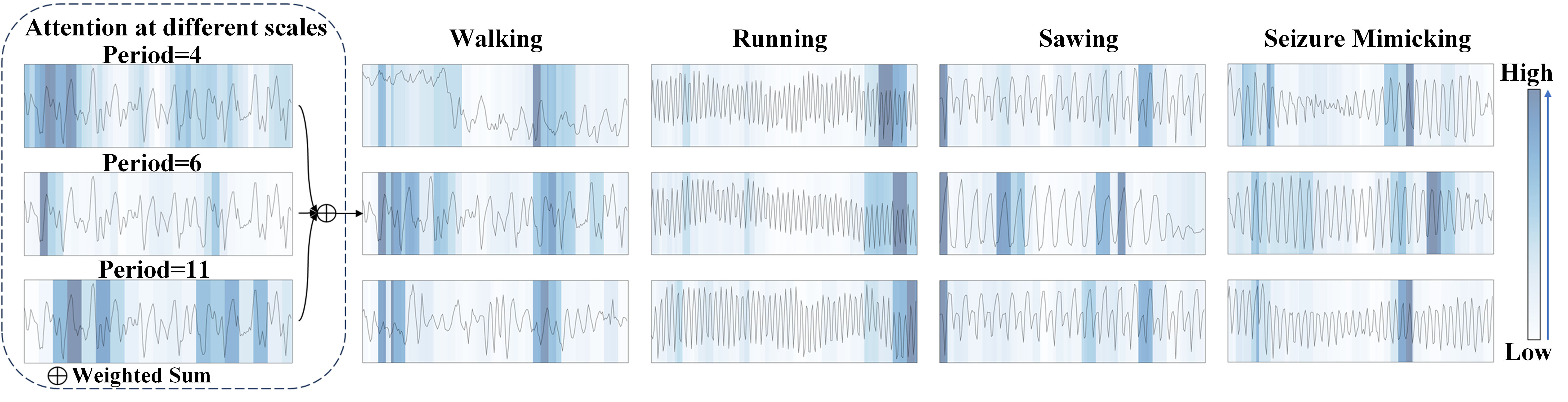}
\caption{Interpretability visualization of the MPTSNet model on the Epilepsy dataset. Left: attention maps at different periodic scales (single example); Right: composite attention maps by weighted summing the individual scale maps. MPTSNet captures varying local patterns across periodic scales while achieving enhanced interpretability by the integration of multi-scale attention.}
\label{interpret}
\end{figure*}

\subsection{Ablation Studies}
\subsubsection{Model Design Variants}
We conducted ablation studies on MPTSNet with three variants across 10 datasets in Setting 1:
\begin{itemize}
    \item \textbf{w/o-LocalExt}: Removes the Local Extractor, capturing only global dependencies at each time point.
    \item \textbf{w/o-GlobalCap}: Removes the Global Capturer, relying solely on local patterns for each segment. 
    \item \textbf{w/o-MP}: Omitted the multi-scale FFT decomposition, applying Local Extractor and Global Capturer to the entire time series instead.
\end{itemize}

The results of our ablation studies shown in Table~\ref{ablation} provide significant insights into the contributions of different components of the MPTSNet model. The full MPTSNet model achieved the highest average accuracy of 0.754, demonstrating the effectiveness of integrating local pattern extraction, global dependency modeling, and multiscale periodic decomposition. Removing the Local Extractor (w/o-LocalExt) and the Global Capturer (w/o-GlobalCap) resulted in reduced accuracies of 0.737 and 0.731, respectively, highlighting their critical roles in capturing local and global features. The most significant performance drop was observed without Multiscale Periodic decomposition (w/o-MP), with accuracy falling to 0.709, underscoring the importance of multiscale periodic feature analysis.

\begin{table}[htbp]
\centering
\setlength{\tabcolsep}{1mm}
\begin{tabular}{lcc|lc}
\toprule
Model Design Variants & Accuracy &  & Setting of \textit{k} & Accuracy \\
\midrule
MPTSNet & \textbf{0.754} &  & \textit{k}=7 & 0.751 \\
w/o-LocalExt & 0.737 &  & \textit{k}=5 & \textbf{0.754} \\
w/o-GlobalCap & 0.731 &  & \textit{k}=3 & 0.742 \\
w/o-MP & 0.709 &  & \textit{k}=1 & 0.734 \\
\bottomrule
\end{tabular}
\caption{The ablation study was conducted on 10 UEA datasets within Setting 1 experiments. Average accuracies are reported, with the best performance indicated in \textit{bold}.}
\label{ablation}
\end{table}

\subsubsection{Setting of \textit{k}}
The hyperparameter \textit{k} is the sole parameter that needs to be set in MPTSNet to determine the number of periodic scales. Experimental results in Table~\ref{ablation} show a consistent improvement in average accuracy as \textit{k} increases from 1 (0.734) to 5 (0.754), followed by a slight decline at \textit{k}=7 (0.751). This trend suggests that incorporating multiple periodic scales enhances the model capacity to capture complex temporal patterns, with \textit{k}=5 providing an optimal balance between information extraction and noise avoidance. The marginal performance decrease at \textit{k}=7 indicates that excessive periodic scale extraction may introduce noise, particularly for datasets with less pronounced periodicity. These findings underscore the importance of judicious \textit{k} selection, balancing the trade-off between capturing sufficient periodic information and mitigating the risk of overfitting. While \textit{k}=5 emerges as the optimal value for the diverse datasets examined, further research into adaptive \textit{k}-selection methods could potentially enhance the generalization capabilities of MPTSNet across varied time series data.

\subsection{Enhanced Interpretability}
Our proposed MPTSNet model offers a unique insight into the decision-making process through its interpretability visualization. By integrating global attention across various periodic scales and weighting them accordingly, we can pinpoint the specific time intervals that contribute most significantly to the classification of each time series. As an example, we illustrate the visualization of the attention maps using one of the datasets, the Epilepsy dataset in Fig.~\ref{interpret}. This dataset, sourced from the UEA archive, encompasses wrist motion recordings from multiple participants engaged in four distinct activities: Walking, Running, Sawing, and Seizure Mimicking. Our analysis is conducted on three randomly selected samples from each class.

The left portion of the figure presents the attention maps from different periodic scales, which clearly demonstrate that the model attention is drawn to varying temporal locations across different periodic scales. To obtain a holistic view, we computed composite attention maps by combining individual scale attention maps, weighted by their corresponding amplitudes. These composite maps, presented in the right panel, are juxtaposed with the original signals. The resulting visualizations reveal distinct patterns for different classes while maintaining similarities within the same class. This observation underscores the ability of MPTSNet to differentiate between various time series classes by identifying and exploiting key temporal patterns and extracting meaningful representations.

\section{Conclusion}
To improve the performance of multivariate time series classification, we introduce MPTSNet, a novel deep learning framework. Our approach incorporates periodicity as a fundamental time scale and leverages the complementary strengths of CNN and attention mechanisms to thoroughly learn multiscale local and global features within time series data. Extensive experiments demonstrate the effectiveness of MPTSNet in various real-world applications. By visualizing attention maps, we provide insights into how MPTSNet identifies crucial temporal patterns, enabling more interpretable and reliable classification results.

\section{Acknowledgments}
The work is jointly supported by the German Federal Ministry of Education and Research (BMBF) in the framework of the international future AI lab "AI4EO -- Artificial Intelligence for Earth Observation: Reasoning, Uncertainties, Ethics and Beyond" (grant number: 01DD20001), by German Federal Ministry for Economic Affairs and Climate Action in the framework of the "national center of excellence ML4Earth" (grant number: 50EE2201C), by the European Union’s Horizon Europe research and innovations actions programme in the framework of the "Multi-source and Multi-scale Earth observation and Novel Machine Learning Methods for Mineral Exploration and Mine Site Monitoring (MultiMiner)" (grant number: 101091374), by the Excellence Strategy of the Federal Government and the Länder through the TUM Innovation Network EarthCare and by Munich Center for Machine Learning.

\bibliography{aaai25}

\begin{thebibliography}{40}
\providecommand{\natexlab}[1]{#1}

\bibitem[{An et~al.(2023)An, Rahman, Zhou, and Kang}]{an2023comprehensive}
An, Q.; Rahman, S.; Zhou, J.; and Kang, J.~J. 2023.
\newblock A comprehensive review on machine learning in healthcare industry: classification, restrictions, opportunities and challenges.
\newblock \emph{Sensors}, 23(9): 4178.

\bibitem[{Bagnall et~al.(2018)Bagnall, Dau, Lines, Flynn, Large, Bostrom, Southam, and Keogh}]{bagnall2018uea}
Bagnall, A.; Dau, H.~A.; Lines, J.; Flynn, M.; Large, J.; Bostrom, A.; Southam, P.; and Keogh, E. 2018.
\newblock The UEA multivariate time series classification archive, 2018.
\newblock \emph{arXiv preprint arXiv:1811.00075}.

\bibitem[{Baydogan, Runger, and Tuv(2013)}]{baydogan2013bag}
Baydogan, M.~G.; Runger, G.; and Tuv, E. 2013.
\newblock A bag-of-features framework to classify time series.
\newblock \emph{IEEE transactions on pattern analysis and machine intelligence}, 35(11): 2796--2802.

\bibitem[{Cai et~al.(2024)Cai, Liang, Liu, Feng, and Wu}]{cai2024msgnet}
Cai, W.; Liang, Y.; Liu, X.; Feng, J.; and Wu, Y. 2024.
\newblock Msgnet: Learning multi-scale inter-series correlations for multivariate time series forecasting.
\newblock In \emph{Proceedings of the AAAI Conference on Artificial Intelligence}, volume~38, 11141--11149.

\bibitem[{Chen et~al.(2024)Chen, Qiu, Zhu, Li, Wang, Sotiras, Wang, and Razi}]{chen2024timemil}
Chen, X.; Qiu, P.; Zhu, W.; Li, H.; Wang, H.; Sotiras, A.; Wang, Y.; and Razi, A. 2024.
\newblock TimeMIL: Advancing Multivariate Time Series Classification via a Time-aware Multiple Instance Learning.
\newblock \emph{arXiv preprint arXiv:2405.03140}.

\bibitem[{Cui, Chen, and Chen(2016)}]{cui2016multi}
Cui, Z.; Chen, W.; and Chen, Y. 2016.
\newblock Multi-scale convolutional neural networks for time series classification.
\newblock \emph{arXiv preprint arXiv:1603.06995}.

\bibitem[{Farahani et~al.(2023)Farahani, McCormick, Gianinny, Hudacheck, Harik, Liu, and Wuest}]{farahani2023time}
Farahani, M.~A.; McCormick, M.; Gianinny, R.; Hudacheck, F.; Harik, R.; Liu, Z.; and Wuest, T. 2023.
\newblock Time-series pattern recognition in Smart Manufacturing Systems: A literature review and ontology.
\newblock \emph{Journal of Manufacturing Systems}, 69: 208--241.

\bibitem[{Gu, Goel, and R{\'e}(2021)}]{gu2021efficiently}
Gu, A.; Goel, K.; and R{\'e}, C. 2021.
\newblock Efficiently modeling long sequences with structured state spaces.
\newblock \emph{arXiv preprint arXiv:2111.00396}.

\bibitem[{Ienco and Gaetano(2007)}]{ienco2007tiselac}
Ienco, D.; and Gaetano, R. 2007.
\newblock Tiselac: time series land cover classification challenge.
\newblock \emph{TiSeLaC: Time Series Land Cover Classification Challenge}, 2.

\bibitem[{Karim et~al.(2017)Karim, Majumdar, Darabi, and Chen}]{karim2017lstm}
Karim, F.; Majumdar, S.; Darabi, H.; and Chen, S. 2017.
\newblock LSTM fully convolutional networks for time series classification.
\newblock \emph{IEEE access}, 6: 1662--1669.

\bibitem[{Karim et~al.(2019)Karim, Majumdar, Darabi, and Harford}]{karim2019multivariate}
Karim, F.; Majumdar, S.; Darabi, H.; and Harford, S. 2019.
\newblock Multivariate LSTM-FCNs for time series classification.
\newblock \emph{Neural networks}, 116: 237--245.

\bibitem[{Lai et~al.(2018)Lai, Chang, Yang, and Liu}]{lai2018modeling}
Lai, G.; Chang, W.-C.; Yang, Y.; and Liu, H. 2018.
\newblock Modeling long-and short-term temporal patterns with deep neural networks.
\newblock In \emph{The 41st international ACM SIGIR conference on research \& development in information retrieval}, 95--104.

\bibitem[{Lai et~al.(2023)Lai, Zhang, Liu, Qian, Wei, Chen, Chen, and Yin}]{lai2023interformer}
Lai, Z.-H.; Zhang, T.-H.; Liu, Q.; Qian, X.; Wei, L.-F.; Chen, S.-L.; Chen, F.; and Yin, X.-C. 2023.
\newblock InterFormer: Interactive Local and Global Features Fusion for Automatic Speech Recognition.
\newblock \emph{arXiv preprint arXiv:2305.16342}.

\bibitem[{Li et~al.(2021)Li, Choi, Xu, Bhowmick, Chun, and Wong}]{li2021shapenet}
Li, G.; Choi, B.; Xu, J.; Bhowmick, S.~S.; Chun, K.-P.; and Wong, G. L.-H. 2021.
\newblock Shapenet: A shapelet-neural network approach for multivariate time series classification.
\newblock In \emph{Proceedings of the AAAI conference on artificial intelligence}, volume~35, 8375--8383.

\bibitem[{Li et~al.(2023{\natexlab{a}})Li, Yang, Su, Li, and Wang}]{li2023human}
Li, Y.; Yang, G.; Su, Z.; Li, S.; and Wang, Y. 2023{\natexlab{a}}.
\newblock Human activity recognition based on multienvironment sensor data.
\newblock \emph{Information Fusion}, 91: 47--63.

\bibitem[{Li et~al.(2023{\natexlab{b}})Li, Rao, Pan, and Xu}]{li2023mts}
Li, Z.; Rao, Z.; Pan, L.; and Xu, Z. 2023{\natexlab{b}}.
\newblock Mts-mixers: Multivariate time series forecasting via factorized temporal and channel mixing.
\newblock \emph{arXiv preprint arXiv:2302.04501}.

\bibitem[{Lines et~al.(2012)Lines, Davis, Hills, and Bagnall}]{lines2012shapelet}
Lines, J.; Davis, L.~M.; Hills, J.; and Bagnall, A. 2012.
\newblock A shapelet transform for time series classification.
\newblock In \emph{Proceedings of the 18th ACM SIGKDD international conference on Knowledge discovery and data mining}, 289--297.

\bibitem[{Liu et~al.(2024)Liu, Yang, Liu, Chen, Liang, Wang, Cui, and Gu}]{liu2024todynet}
Liu, H.; Yang, D.; Liu, X.; Chen, X.; Liang, Z.; Wang, H.; Cui, Y.; and Gu, J. 2024.
\newblock Todynet: temporal dynamic graph neural network for multivariate time series classification.
\newblock \emph{Information Sciences}, 120914.

\bibitem[{Liu et~al.(2022)Liu, Zeng, Chen, Xu, Lai, Ma, and Xu}]{liu2022scinet}
Liu, M.; Zeng, A.; Chen, M.; Xu, Z.; Lai, Q.; Ma, L.; and Xu, Q. 2022.
\newblock Scinet: Time series modeling and forecasting with sample convolution and interaction.
\newblock \emph{Advances in Neural Information Processing Systems}, 35: 5816--5828.

\bibitem[{Luo and Wang(2024)}]{luo2024moderntcn}
Luo, D.; and Wang, X. 2024.
\newblock Moderntcn: A modern pure convolution structure for general time series analysis.
\newblock In \emph{The Twelfth International Conference on Learning Representations}.

\bibitem[{Nie et~al.(2023)Nie, Nguyen, Sinthong, and Kalagnanam}]{nie2023a}
Nie, Y.; Nguyen, N.~H.; Sinthong, P.; and Kalagnanam, J. 2023.
\newblock A Time Series is Worth 64 Words: Long-term Forecasting with Transformers.
\newblock In \emph{The Eleventh International Conference on Learning Representations}.

\bibitem[{Ru{\ss}wurm et~al.(2023)Ru{\ss}wurm, Courty, Emonet, Lef{\`e}vre, Tuia, and Tavenard}]{russwurm2023end}
Ru{\ss}wurm, M.; Courty, N.; Emonet, R.; Lef{\`e}vre, S.; Tuia, D.; and Tavenard, R. 2023.
\newblock End-to-end learned early classification of time series for in-season crop type mapping.
\newblock \emph{ISPRS Journal of Photogrammetry and Remote Sensing}, 196: 445--456.

\bibitem[{Sch{\"a}fer and Leser(2017)}]{schafer2017multivariate}
Sch{\"a}fer, P.; and Leser, U. 2017.
\newblock Multivariate time series classification with WEASEL+ MUSE.
\newblock \emph{arXiv preprint arXiv:1711.11343}.

\bibitem[{Tang et~al.(2022)Tang, Long, Liu, Zhou, Blumenstein, and Jiang}]{tangomni}
Tang, W.; Long, G.; Liu, L.; Zhou, T.; Blumenstein, M.; and Jiang, J. 2022.
\newblock Omni-Scale CNNs: a simple and effective kernel size configuration for time series classification.
\newblock In \emph{International Conference on Learning Representations}.

\bibitem[{Wang et~al.(2023)Wang, Peng, Huang, Wang, Chen, and Xiao}]{wang2023micn}
Wang, H.; Peng, J.; Huang, F.; Wang, J.; Chen, J.; and Xiao, Y. 2023.
\newblock Micn: Multi-scale local and global context modeling for long-term series forecasting.
\newblock In \emph{The eleventh international conference on learning representations}.

\bibitem[{Wang et~al.(2022)Wang, Chen, Hershkovich, Yang, Shetty, Singh, Jiang, Kotla, Shang, Yerrabelli et~al.}]{wang2022systematic}
Wang, W.~K.; Chen, I.; Hershkovich, L.; Yang, J.; Shetty, A.; Singh, G.; Jiang, Y.; Kotla, A.; Shang, J.~Z.; Yerrabelli, R.; et~al. 2022.
\newblock A systematic review of time series classification techniques used in biomedical applications.
\newblock \emph{Sensors}, 22(20): 8016.

\bibitem[{Wen et~al.(2022)Wen, Zhou, Zhang, Chen, Ma, Yan, and Sun}]{wen2022transformers}
Wen, Q.; Zhou, T.; Zhang, C.; Chen, W.; Ma, Z.; Yan, J.; and Sun, L. 2022.
\newblock Transformers in time series: A survey.
\newblock \emph{arXiv preprint arXiv:2202.07125}.

\bibitem[{Wu et~al.(2023)Wu, Hu, Liu, Zhou, Wang, and Long}]{wu2023timesnet}
Wu, H.; Hu, T.; Liu, Y.; Zhou, H.; Wang, J.; and Long, M. 2023.
\newblock TimesNet: Temporal 2D-Variation Modeling for General Time Series Analysis.
\newblock In \emph{The Eleventh International Conference on Learning Representations}.

\bibitem[{Wu et~al.(2022)Wu, Wu, Xu, Wang, and Long}]{wu2022flowformer}
Wu, H.; Wu, J.; Xu, J.; Wang, J.; and Long, M. 2022.
\newblock Flowformer: Linearizing transformers with conservation flows.
\newblock \emph{arXiv preprint arXiv:2202.06258}.

\bibitem[{Yang et~al.(2015)Yang, Nguyen, San, Li, and Krishnaswamy}]{yang2015deep}
Yang, J.; Nguyen, M.~N.; San, P.~P.; Li, X.; and Krishnaswamy, S. 2015.
\newblock Deep convolutional neural networks on multichannel time series for human activity recognition.
\newblock In \emph{Ijcai}, volume~15, 3995--4001. Buenos Aires, Argentina.

\bibitem[{Yu, Kim, and Mechefske(2021)}]{yu2021analysis}
Yu, W.; Kim, I.~Y.; and Mechefske, C. 2021.
\newblock Analysis of different RNN autoencoder variants for time series classification and machine prognostics.
\newblock \emph{Mechanical Systems and Signal Processing}, 149: 107322.

\bibitem[{Zeng et~al.(2023)Zeng, Chen, Zhang, and Xu}]{zeng2023transformers}
Zeng, A.; Chen, M.; Zhang, L.; and Xu, Q. 2023.
\newblock Are transformers effective for time series forecasting?
\newblock In \emph{Proceedings of the AAAI conference on artificial intelligence}, volume~37, 11121--11128.

\bibitem[{Zhang et~al.(2022)Zhang, Zhang, Cao, Bian, Yi, Zheng, and Li}]{zhang2022less}
Zhang, T.; Zhang, Y.; Cao, W.; Bian, J.; Yi, X.; Zheng, S.; and Li, J. 2022.
\newblock Less is more: Fast multivariate time series forecasting with light sampling-oriented mlp structures.
\newblock \emph{arXiv preprint arXiv:2207.01186}.

\bibitem[{Zhang et~al.(2020)Zhang, Gao, Lin, and Lu}]{zhang2020tapnet}
Zhang, X.; Gao, Y.; Lin, J.; and Lu, C.-T. 2020.
\newblock Tapnet: Multivariate time series classification with attentional prototypical network.
\newblock In \emph{Proceedings of the AAAI conference on artificial intelligence}, volume~34, 6845--6852.

\bibitem[{Zhang and Yan(2023)}]{zhang2023crossformer}
Zhang, Y.; and Yan, J. 2023.
\newblock Crossformer: Transformer utilizing cross-dimension dependency for multivariate time series forecasting.
\newblock In \emph{The eleventh international conference on learning representations}.

\bibitem[{Zhao et~al.(2017)Zhao, Lu, Chen, Liu, and Wu}]{zhao2017convolutional}
Zhao, B.; Lu, H.; Chen, S.; Liu, J.; and Wu, D. 2017.
\newblock Convolutional neural networks for time series classification.
\newblock \emph{Journal of systems engineering and electronics}, 28(1): 162--169.

\bibitem[{Zhao et~al.(2024)Zhao, Li, Hong, and Shen}]{zhao2024metarocketc}
Zhao, J.; Li, Q.; Hong, Y.; and Shen, M. 2024.
\newblock MetaRockETC: Adaptive Encrypted Traffic Classification in Complex Network Environments via Time Series Analysis and Meta-Learning.
\newblock \emph{IEEE Transactions on Network and Service Management}.

\bibitem[{Zheng et~al.(2014)Zheng, Liu, Chen, Ge, and Zhao}]{zheng2014time}
Zheng, Y.; Liu, Q.; Chen, E.; Ge, Y.; and Zhao, J.~L. 2014.
\newblock Time series classification using multi-channels deep convolutional neural networks.
\newblock In \emph{International conference on web-age information management}, 298--310. Springer.

\bibitem[{Zhou et~al.(2022)Zhou, Ma, Wen, Wang, Sun, and Jin}]{zhou2022fedformer}
Zhou, T.; Ma, Z.; Wen, Q.; Wang, X.; Sun, L.; and Jin, R. 2022.
\newblock Fedformer: Frequency enhanced decomposed transformer for long-term series forecasting.
\newblock In \emph{International conference on machine learning}, 27268--27286. PMLR.

\bibitem[{Zuo et~al.(2023)Zuo, Li, Choi, Bhowmick, Mah, and Wong}]{zuo2023svp}
Zuo, R.; Li, G.; Choi, B.; Bhowmick, S.~S.; Mah, D. N.-y.; and Wong, G.~L. 2023.
\newblock SVP-T: a shape-level variable-position transformer for multivariate time series classification.
\newblock In \emph{Proceedings of the AAAI Conference on Artificial Intelligence}, volume~37, 11497--11505.

\end{thebibliography}
\end{document}